%% file: root.tex
\documentclass[letterpaper, 10 pt, conference]{ieeeconf}  
\IEEEoverridecommandlockouts                              
\usepackage{epsfig} 
\usepackage{amsmath} 
\usepackage{amssymb}  
\usepackage{color}
\usepackage{cite}
\usepackage{txfonts}
\usepackage{graphicx}
\usepackage{subcaption}
\usepackage[linesnumbered,ruled]{algorithm2e}
\usepackage[normalem]{ulem} 

\makeatletter
\let\NAT@parse\undefined
\makeatother

\usepackage{booktabs}
\usepackage{arydshln}
\usepackage{multirow}
\usepackage{mathrsfs}
\usepackage{soul}

\usepackage{xcolor} 
\usepackage{pifont}   
\usepackage[absolute,overlay]{textpos}
\usepackage{multicol}
\usepackage{colortbl}
\usepackage{makecell}
\usepackage{balance}
\usepackage{cancel}
\usepackage{adjustbox}

\usepackage{url}
\usepackage{hyperref}
\hypersetup{
    colorlinks = true, 
    linkcolor = red,
    citecolor = green
}

\title{\LARGE \bf
ComDrive: Comfort-Oriented End-to-End Autonomous Driving
}

\author{Junming Wang$^{1,2,*}$, 
Xingyu Zhang$^{1,*}$, 
Zebin Xing$^{1,3}$,
Songen Gu$^{1,3}$, \\
Xiaoyang Guo$^{1}$, 
Yang Hu$^{1}$, 
Ziying Song$^{1,5}$, 
Qian Zhang$^{1}$, 
Xiaoxiao Long$^{4}$,
Wei Yin$^{1,\dagger}$ \\
\\ 
\thanks{ $^{*}$Equal Contribution. $^{\dag}$Corresponding Author. }   
\thanks{$^{1}$Horizon Robotics. $^{2}$University of Hong Kong. $^3$University of the Chinese Academy of Sciences. $^{4}$Nanjing University. $^5$Beijing Jiaotong University.} 
}

\begin{document}

\maketitle
\thispagestyle{empty}
\pagestyle{empty}

\begin{abstract}
We propose \textit{\textbf{ComDrive}}: the first comfort-oriented end-to-end autonomous driving system to generate temporally consistent and comfortable trajectories. Recent studies have demonstrated that imitation learning-based planners and learning-based trajectory scorers can effectively generate and select safety trajectories that closely mimic expert demonstrations. However, such trajectory planners and scorers face the challenge of generating temporally inconsistent and uncomfortable trajectories. To address these issues, ComDrive first extracts 3D spatial representations through sparse perception, which then serves as conditional inputs. These inputs are used by a Conditional Denoising Diffusion Probabilistic Model (DDPM)-based motion planner to generate temporally consistent multi-modal trajectories. A dual-stream adaptive trajectory scorer subsequently selects the most comfortable trajectory from these candidates to control the vehicle. Experiments demonstrate that ComDrive achieves state-of-the-art performance in both comfort and safety, outperforming UniAD by \textbf{17\%}in driving comfort and reducing collision rates by \textbf{25\%}compared to SparseDrive. More results are available on our project page: \url{https://jmwang0117.github.io/ComDrive/}.
\end{abstract}

\section{Introduction}
\label{sec:intro}

Recent advancements in autonomous driving technology have focused on safety-oriented end-to-end paradigms \cite{uniad,vad, sun2024sparsedrive,shao2023reasonnet,li2024ego}. These methods integrate perception, planning, and trajectory scoring into unified models, aiming to mitigate collision risks in complex traffic scenarios. The latest research proposes imitation learning-based motion planners \cite{chen2024vadv2,cheng2024pluto} that learn driving strategies from large-scale driving demonstrations and employ learning-based trajectory scorers \cite{zhao2021tnt,jiang2023vad} to select the safest trajectory from multiple predicted candidates to control the vehicle.  

Unfortunately, despite significant advancements in prediction accuracy and safety, these systems continue to face the dilemma of an \textbf{\textit{uncomfortable}} riding experience \cite{shao2023safety,wang2021socially}. This degradation in comfort can be attributed to two primary factors: 1) temporally inconsistent trajectory generation, i.e., unstable and non-smooth predictions across consecutive time steps; and 2) the inability of trajectory scorers to adaptively update their strategies in response to changing environmental conditions. This lack of adaptability often leads to the selection of suboptimal trajectories that result in continuous braking or excessive turning curvature. These two issues significantly compromise the overall comfort of autonomous vehicle passengers.

\begin{figure}[t]
  \centering
  \begin{subfigure}{\linewidth}
    \centering
    \includegraphics[width=\textwidth]{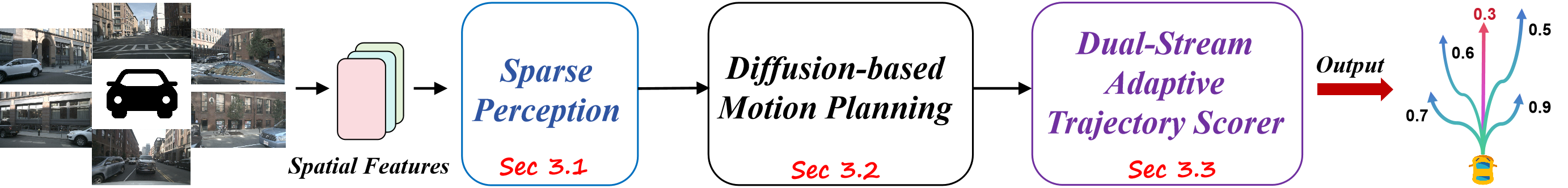}  
    \caption{Our End-to-End Autonomous Driving Paradigm}
    \label{fig:paradigm}
  \end{subfigure}%
  
  \vspace{5pt} 
  
  \begin{subfigure}{\linewidth}
    \centering
    \includegraphics[width=\textwidth]{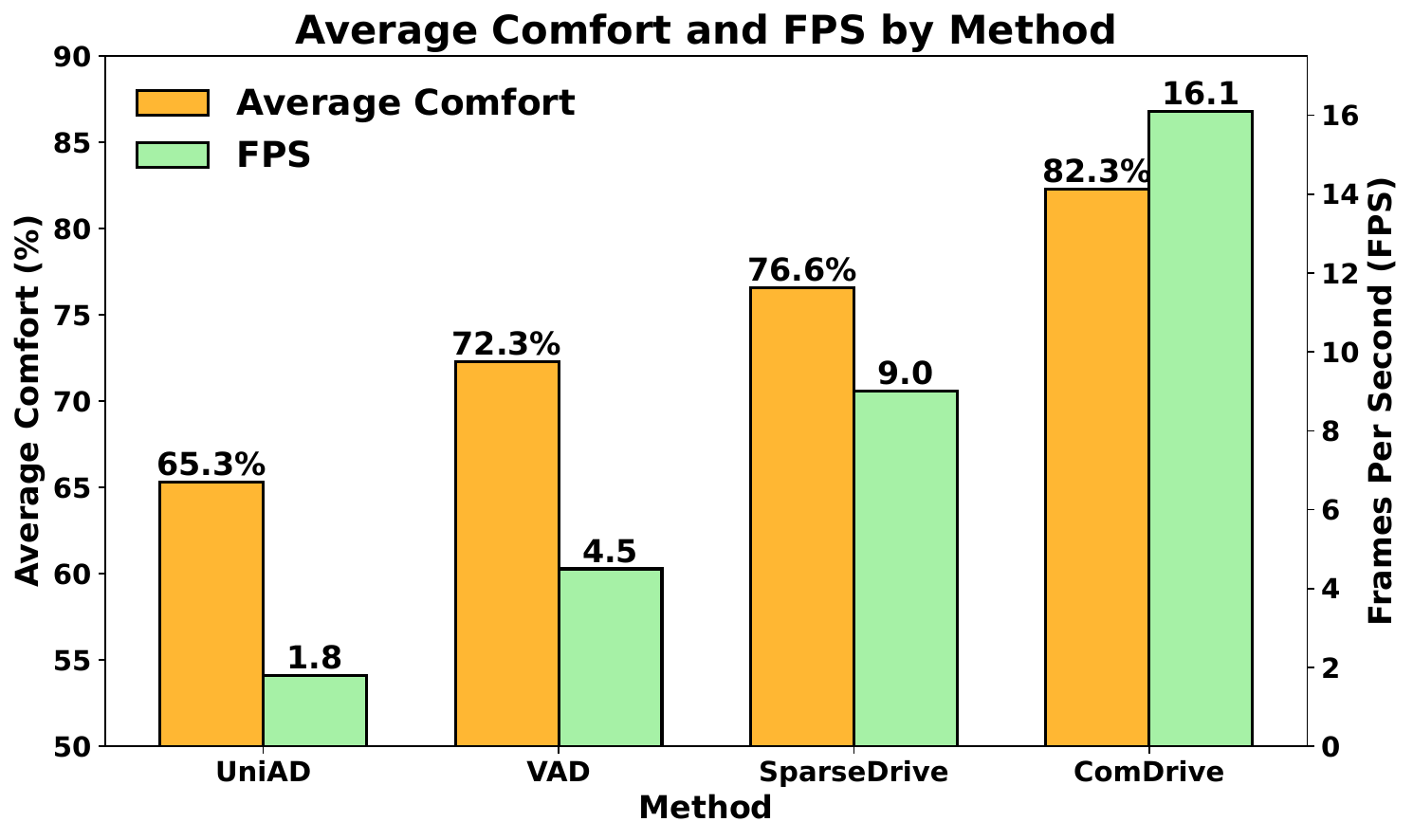}  
    \caption{Performance Comparison of Comfort and Efficiency}
    \label{fig:comfort-metrics}
  \end{subfigure}

  \caption{\small \textbf{Architecture and Performance Evaluation of ComDrive.}}
  \label{fig:head}
\end{figure}

In this work, we introduce \textit{\textbf{ComDrive}}, the first comfort-oriented end-to-end autonomous driving system (in Fig. \ref{fig:head}) designed to address the aforementioned challenges. We identify that the \textbf{\textit{temporal inconsistency}} in trajectories generated by imitation learning-based planners stems from two primary factors: inadequate temporal correlation and limited generalization capability. Firstly, these planners typically rely solely on current frame information to forecast future multi-modal trajectories, neglecting the crucial temporal correlations between consecutive predictions \cite{zhou2023query,tang2024hpnet}. Secondly, their performance is inherently constrained by the quality of collected offline expert trajectories, rendering them vulnerable to changes in system dynamics and out-of-distribution states. Consequently, the learned policies often lack robust generalization to novel scenarios. Drawing inspiration from the diffusion policy \cite{chi2024diffusionpolicy}, which gracefully learns multimodal action distributions in robotic manipulation, we propose an innovative diffusion-based motion planner. This planner is designed to generate multimodal trajectories with strong temporal consistency.

Moreover, the suboptimal trajectory selection stems from the trajectory scorer's limitations and the absence of a universal comfort metric. Recent studies have revealed that learning-based scorers are inferior to rule-based scorers in closed-loop scenarios \cite{dauner2023parting}, while the latter suffers from limited generalization due to their reliance on hand-crafted post-processing. Other researchers have explored the use of Vision-Language Models (VLMs) \cite{shao2024lmdrive,sima2023drivelm,mao2023gpt} to perceive the motion of surrounding agents to decide the next movement. However, directly employing VLMs as driving decision-makers poses challenges related to poor interpretability and severe hallucinations \cite{xu2024hallucination}. To address these issues, we propose a novel dual-stream adaptive trajectory scorer and universal comfort metric (Fig. \ref{fig:head}) that combines the interpretability of rule-based scorers with the adaptability of VLMs to dynamically adjust driving styles (\textit{i.e., aggressive or conservative}) for continuous evaluation.

In summary, ComDrive first utilizes sparse perception to detect, track, and map driving scenarios based on sparse features, generating 3D spatial representations. These representations are then conditionally fed into a diffusion-based motion planner, powered by a Conditional Denoising Diffusion Probabilistic Model (DDPM), which generates multiple candidate trajectories. Subsequently, a dual-stream adaptive trajectory scorer evaluates these candidates, combining rule-based scoring with VLM-guided (\textit{i.e., Llama 3.2V}) driving style assessment to select the most comfortable and safe trajectory for vehicle control. A key feature of this scoring system lies in the VLM component's ability to infer the appropriate driving style for the current context and dynamically adjust the weights of the rule-based scorer. This adaptive mechanism ensures context-aware trajectory selection, enhancing the system's capacity to balance safety and comfort in different scenarios. The main contributions of our work are summarized as follows:
\begin{itemize}
\item \textbf{Diffusion-based Motion Planner:} We propose a novel diffusion-based motion planner that generates temporal consistent and multi-modal trajectories by conditioning on the 3D representation extracted by the sparse perception network and incorporating the speed, acceleration, and yaw of the historical prediction trajectory. (§~\ref{sec:Diffusion})

\item \textbf{Plug-and-Play Trajectory Scorer:}  We introduce a novel dual-stream adaptive trajectory scorer (DATS) and a comfort metric, which address the gap in comfort-oriented driving, making it easily integrated into existing autonomous driving systems. (§~\ref{sec:Scorer})

\item \textbf{Excellent Results on Public Benchmarks:} ComDrive achieves state-of-the-art performance (\textit{i.e.}, reduces the average collision rate by \textbf{71\%} compared to VAD) and efficiency (\textit{i.e.}, \textbf{1.9$\times$} faster than SparseDrive) on nuScenes, while increasing comfort by \textbf{32\%} on real-world datasets, showcasing its effectiveness across various scenarios. (§~\ref{sec:nuScenes} and §~\ref{sec:real-world})
\end{itemize}

\section{Related work}
\label{sec:related_work}

\begin{figure*}[t]
  \centering
     \includegraphics[width=\linewidth]{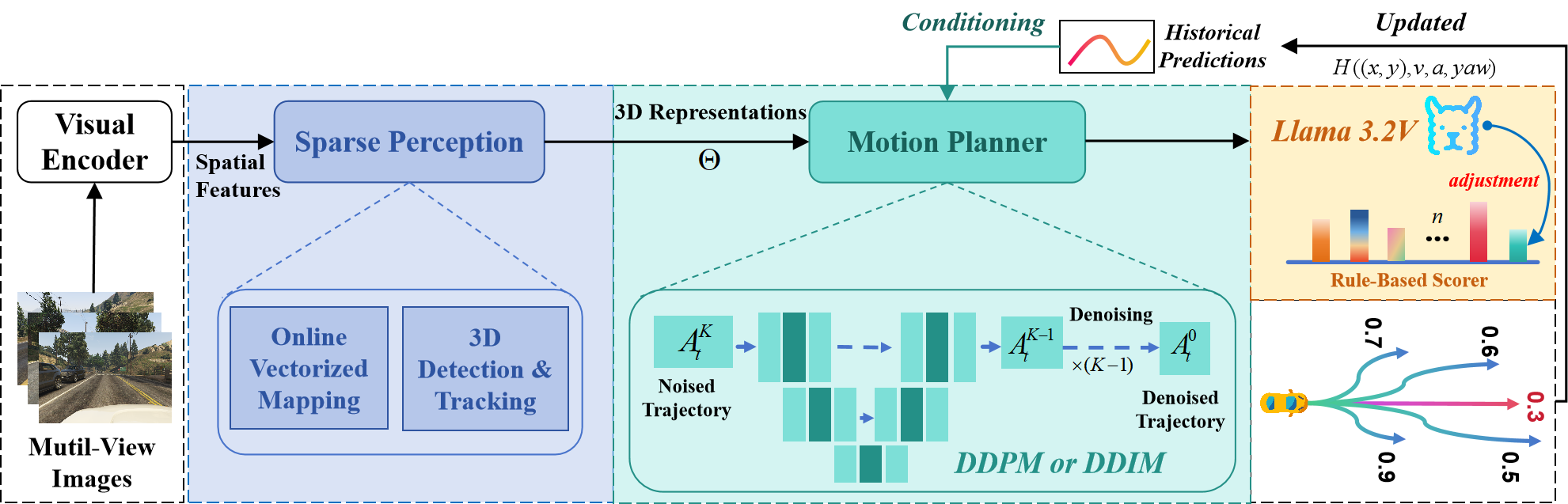}
   \caption{\small \textbf{Overview of our proposed framework.} ComDrive first extracts features from multi-view images using an off-the-shelf visual encoder then perceives dynamic and static elements sparsely to generate 3D representation. The above representations and historical prediction trajectories are used as conditions of the diffusion model to generate \textbf{\textit{temporal consistency}} multi-modal trajectories. The final trajectory scorer selects the most \textbf{\textit{comfortable}} trajectory from these candidates to control the vehicle. }
   \label{fig:overview}
\end{figure*}

\subsection{End-to-End Autonomous Driving}
\label{subsec:e2e_autonomous_driving}

End-to-end autonomous driving \cite{shao2024lmdrive,li2024hydra,zheng2024genad,doll2024dualad} aims to generate planning trajectories directly from raw sensors. In the field, advancements have been categorized based on their evaluation methods: open-loop and closed-loop systems. In open-loop systems, UniAD \cite{hu2023planning} presents a unified framework that integrates full-stack driving tasks with query-unified interfaces for improved interaction between tasks. VAD \cite{jiang2023vad} boosts planning safety and efficiency, evidenced by its performance on the nuScenes dataset, while SparseDrive \cite{sun2024sparsedrive} utilizes sparse representations to mitigate information loss and error propagation inherent in modular systems, enhancing both task performance and computational efficiency. For closed-loop evaluations, VADv2 \cite{chen2024vadv2} advances vectorized autonomous driving with probabilistic planning, using multi-view images to generate action distributions for vehicle control, excelling in the CARLA Town05 benchmark. 

\subsection{Diffusion Models for Trajectory Generation}
\label{subsec:diffusion_models}

Diffusion models initially celebrated in image synthesis, have been adeptly repurposed for trajectory generation \cite{pearce2023imitating,reuss2023goal,janner2022planning}. Potential-Based Diffusion Motion Planning \cite{luo2024potential} further enhances the field by employing learned potential functions to construct adaptable motion plans for cluttered environments. NoMaD \cite{sridhar2024nomad} and SkillDiffuser \cite{liang2024skilldiffuser} both present unified frameworks that streamline goal-oriented navigation and skill-based task execution, respectively, with NoMaD achieving improved navigation outcomes and SkillDiffuser enabling interpretable, high-level instruction following. In a word, diffusion models offer a promising alternative to imitation learning-based end-to-end autonomous driving frameworks for planning.

\subsection{Large Language Models (LLMs) for Trajectory Evaluation}
\label{subsec:llms_for_trajectory}

Trajectory scoring \cite{fan2018baidu} plays a vital role in autonomous driving decision-making. Rule-based methods \cite{treiber2000congested} provide strong safety guarantees but lack flexibility, while learning-based methods \cite{chitta2021neat} perform well in open-loop tasks but struggle in closed-loop scenarios \cite{treiber2000congested}. Recently, DriveLM \cite{sima2023drivelm} integrates VLMs into end-to-end driving systems, modelling graph-structured reasoning through perception, prediction, and planning question-answer pairs. However, the generated results of large models may contain hallucinations and require further strategies for safe application in autonomous driving. The emergence of VLMs raises the question: \textit{Can VLMs adaptively adjust driving style while ensuring comfort based on a trajectory scorer?}

\section{Methodology}
\label{method}

In this section, we introduce the key components of \textbf{\textit{ComDrive}} (Fig. \ref{fig:overview}): sparse perception (Sec \ref{sec:Sparse}), diffusion-based motion planner (Sec \ref{sec:Diffusion}), and dual-stream adaptive trajectory scorer (Sec \ref{sec:Scorer}).

\subsection{Sparse Perception}
\label{sec:Sparse}

ComDrive begins by employing a visual encoder \cite{he2016deep} to extract visual features, denoted as $\mathcal{F}$, from the input multi-view camera images. These images denoted as $\Gamma = \{J_{\tau} \in \mathbb{R}^{N \times 3 \times H \times W}\}_{{\tau}=T-k}^{T}$, where $N$ is the number of camera views, $k$ is the temporal window length, and $J_{\tau}$ represents the multi-view images at timestep $\tau$, with $T$ being the current timestep. Subsequently, the sparse perception from \cite{sun2024sparsedrive} performs detection, tracking, and online mapping tasks concurrently offering a more efficient and compact 3D representation $\Theta $ of the surrounding environment (in Fig. \ref{fig:overview}). This 3D representation encompasses implicit features of surrounding agents and the map, which is crucial for guiding the subsequent diffusion-based motion planner to generate safe multi-modal trajectories, as obstacle information is embedded within the representation.

\subsection{Diffusion-based Motion Planner}
\label{sec:Diffusion}

Fig. \ref{fig:overview} illustrates the overall pipeline of our diffusion-based motion planner. We adopt a CNN-based diffusion policy \cite{chi2024diffusionpolicy} as the foundation, implementing a conditional DDPM to generate multi-modal trajectories.

\noindent\textbf{Motion Planner Diffusion Policy:} Our motion planner takes as input a set of conditions including a compact 3D representation $\Theta$, historical predicted trajectories $\mathcal{H}$, and their corresponding velocity $v_i$, acceleration $a_i$, and yaw encoding $\theta_i$. These conditions are concatenated to form $C$, which is then injected into every convolutional network layer using FiLM \cite{perez2018film}, providing channel-wise conditioning that guides the trajectory generation from the ego position to the anchor positions. The denoising process begins with Gaussian noise $\mathbf{A}_t^{k}$ of shape $[B, N_a, T_i, P]$, where $B$ is the batch size, $N_a$ is the number of anchors, $T_i$ represents the interval times (0.5s, 1s, 1.5s, 2s, 2.5s, 3s) between navigation points, and $P$ denotes the 2D position $(x, y)$ at each interval. $N_a$ anchors represent multiple possible endpoint positions for the trajectory, enabling the generation of diverse, multi-modal paths. During training, these anchors are created by adding random noise to a single expert trajectory, while in inference, they are initialized as pure random noise. This approach allows the model to learn a distribution of possible trajectories rather than a single deterministic path. Through $k$ iterations, the noisy data is refined into a noise-free 3s future multi-modal trajectory $\mathbf{A}_0$ using the denoising network $\epsilon_\theta$. Each trajectory $\tau_i$ is represented as a set of waypoints $\{(x_t, y_t)\}_{t=1}^{T_i}$. The reverse process is described by:

\begin{equation}
\mathbf{A}_t^{k-1} = \alpha (\mathbf{A}_t^{k} - \gamma \epsilon_\theta (\mathbf{A}_t^{k}, k, \Theta, \mathcal{H})) + \mathcal{N}(0, \sigma^2 I)
\end{equation} where $\alpha$ and $\gamma$ are scaling factors, and $\mathcal{N}(0, \sigma^2 I)$ represents Gaussian noise with mean 0 and variance $\sigma^2$. The incorporation of historical trajectories $\mathcal{H}$ as part of the input conditions plays a vital role in enhancing the temporal consistency and smoothness of the generated trajectories. This approach is designed to match real-world human driving behaviour, where drivers naturally consider their recent movements and the evolving traffic situation to make smooth and predictable decisions. By providing the model with historical trajectory information, we enable it to learn the underlying patterns and dynamics of motion, including subtle changes in velocity, acceleration, and direction. This historical context allows the model to generate trajectories that are not only plausible given the current environment (as captured by $\Theta$) but also consistent with the vehicle's recent motion history. The model can thus better capture the continuity of motion, leading to smoother transitions between past and future trajectories. 

During inference, we employ DDIM \cite{songdenoising} as the noise scheduler, enabling real-time trajectory generation with only 10 reasoning steps while maintaining quality. 


\begin{figure*}[t]
  \centering
     \includegraphics[width=\linewidth]{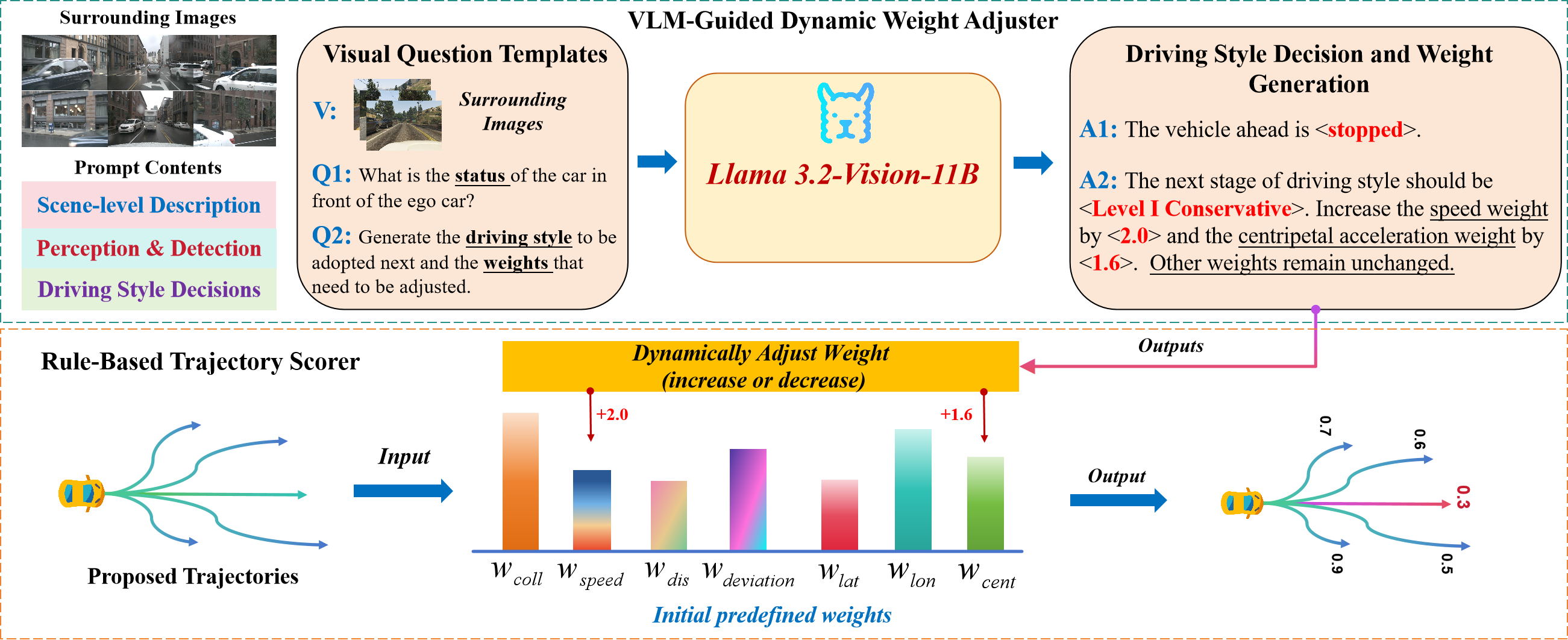}
   \caption{\textbf{Overview of the Dual-Stream Adaptive Trajectory Scorer (DATS).} The system integrates a Rule-Based Scorer with a VLM-Guided Dynamic Weight Adjuster for adaptive and interpretable trajectory scoring.}
   \label{fig:scoring}
\end{figure*}

\subsection{Dual-Stream Adaptive Trajectory Scorer}
\label{sec:Scorer}

To select the comfortable and safe trajectory from the multi-modal paths generated by DDIM, we propose the dual-stream adaptive trajectory scorer (DATS), which consists of two parallel components: the rule-based trajectory scorer and the VLM-guided dynamic weight adjuster, as illustrated in Fig. \ref{fig:scoring}.  

\input{Tables/cost}

\subsubsection{Rule-Based Trajectory Scorer}

We employ a comprehensive scoring strategy that combines safety and comfort considerations (in Table \ref{tab:combined_weights}). The total cost function, \(C_{\text{total}}\), is defined as:
\begin{equation}
C_{\text{total}} = C_{\text{safety}} + C_{\text{comfort}}
\end{equation}

\noindent\textbf{Safety Cost:} The safety cost, $C_{\text{safety}}$, addresses crucial aspects of safe driving:
\begin{equation}
\begin{aligned}
C_{\text{safety}} = & \; w_{\text{coll}} C_{\text{coll}} + w_{\text{dis}} C_{\text{dis}} \\ 
                   & + w_{\text{deviation}} C_{\text{deviation}} + w_{\text{speed}} C_{\text{speed}}
\end{aligned}
\end{equation} Where:
\begin{equation}
C_{\text{coll}} = \begin{cases} 
      1 & \text{if collision detected} \\
      0 & \text{otherwise}
   \end{cases}
\end{equation}
\begin{equation}
C_{\text{dis}} = \lVert\mathbf{p}_{\text{end}} - \mathbf{p}_{\text{target}}\rVert_2
\end{equation}
\begin{equation}
C_{\text{deviation}} = \text{mean}(\arccos(\frac{\mathbf{v}_{\text{target}} \cdot \mathbf{v}_{\text{trajectory}}}{\lVert\mathbf{v}_{\text{target}}\rVert \lVert\mathbf{v}_{\text{trajectory}}\rVert}))
\end{equation}
\begin{equation}
C_{\text{speed}} = \begin{cases} 
      \frac{v_{\text{min}} - \bar{v}}{v_{\text{min}}} & \text{if } \bar{v} < v_{\text{min}} \text{ and style is aggressive} \\
      \frac{\bar{v} - v_{\text{max}}}{v_{\text{max}}} & \text{if } \bar{v} > v_{\text{max}} \text{ and style is conservative} \\
      0 & \text{otherwise}
   \end{cases}
\end{equation}

\noindent\textbf{Comfort Cost:} The comfort cost, $C_{\text{comfort}}$, addresses aspects of driving comfort:
\begin{equation}
C_{\text{comfort}} = w_{\text{lat}} C_{\text{lat}} + w_{\text{lon}} C_{\text{lon}} + w_{\text{cent}} C_{\text{cent}}
\end{equation} Where:
\begin{equation}
C_{\text{lat}} = \max(|l''s^2 + l's''|)
\end{equation}
\begin{equation}
\begin{aligned}
C_{\text{lon}} &= \frac{\sum_{i} (j_i/j_{\text{max}})^2}{\sum_{i} |j_i/j_{\text{max}}| + \epsilon},
C_{\text{cent}} &= \frac{\sum_{i} a_{c,i}^2}{\sum_{i} |a_{c,i}| + \epsilon}
\end{aligned}
\end{equation}

Here, $C_{\text{coll}}$ represents collision risk, penalizing trajectories that collide with obstacles. $C_{\text{dis}}$ measures the distance from the endpoint of the trajectory to the target position. $C_{\text{deviation}}$ evaluates the mean angle deviation between the trajectory and the vector to the target point. $C_{\text{speed}}$ assesses speed appropriateness, ensuring the vehicle maintains a suitable velocity for the driving context and style. In the comfort costs, $C_{\text{lat}}$ penalizes lateral discomfort, $C_{\text{lon}}$ accounts for longitudinal jerk, and $C_{\text{cent}}$ ensures smooth navigation through turns by considering centripetal acceleration. The weights $w_i$ balance these sub-costs (see Table \ref{tab:combined_weights}), allowing the trajectory planner to optimize both safety and comfort based on specific driving requirements. The final selected trajectory is the one with the minimum total cost. 

\subsubsection{VLM-Guided Dynamic Weight Adjuster}

To enhance our rule-based scorer's generalization, we introduce a VLM-guided approach using Llama 3.2 Vision 11B for dynamic weight adjustment. This plug-and-play method, requiring no fine-tuning, interprets complex driving scenarios to inform driving style decisions and weight adjustments.

In the first stage, we create a curated dataset of annotated surround images paired with detailed prompts. These prompts provide comprehensive information about driving scenes, including environmental conditions, agent behaviours, and recommended driving styles with corresponding weight adjustments. This dataset serves as a foundation for priming the VLM, effectively grounding its responses in relevant driving examples. By providing the model with a rich context of driving scenarios and appropriate responses, we aim to reduce model hallucinations. This approach leverages the concept of in-context learning, where the model adapts its behaviour based on the examples it is presented with, without the need for fine-tuning \cite{jin2022good,han2023efficient}. The pre-defined prompts act as a form of implicit knowledge injection, guiding the model to focus on relevant features and appropriate responses in driving scenarios.

The second stage involves the application of this primed knowledge for dynamic weight adjustment. Initially, we use the curated dataset from the first stage to prompt Llama 3.2V, establishing a baseline understanding of driving contexts and appropriate weight adjustments. Subsequently, we implement a periodic activation mechanism for visual question answering (VQA). Using GPT-4o-generated prompt templates, we activate Llama 3.2V at five-second intervals to reassess the driving context and dynamically adjust the weights of our rule-based scoring system (in Fig. \ref{fig:scoring}). 

The five-second interval for VLM activation is carefully chosen based on multiple considerations. Primarily, it reflects the inherent stability of driving scenarios and styles, which typically do not undergo abrupt changes. This temporal consistency in driving contexts allows for a balance between computational efficiency and the need for timely updates. The five-second window is sufficient to capture meaningful changes in the driving environment while aligning with the model's inference latency, ensuring that weight adjustments are both relevant and computationally feasible. While exceptional cases requiring more rapid reassessment may exist, our rule-based method prioritizes safety, mitigating risks associated with less frequent VLM activations. 

It's crucial to note that the VLM's role is to assess driving styles and propose controlled adjustments (Table \ref{tab:combined_weights}) to safety and comfort weights, rather than directly making driving decisions \cite{shao2024lmdrive,sima2023drivelm}. Our approach maintains high safety standards through rule-based mechanisms while leveraging the VLM to achieve comfort-oriented trajectory selection through nuanced adjustments. Additionally, to mitigate cumulative errors, each dynamic update resets weights to their initial states before applying new adjustments, ensuring that each modification is based solely on the current driving context. This reset-and-adjust mechanism enhances the system's robustness and reliability over extended operational periods.

\subsection{End-to-End Driving Comfort Metric}

To address the lack of a universal comfort evaluation metric in existing methods, we propose a general metric to assess the comfort of predicted trajectories \cite{han2023efficient}.  Considering the simplified kinematic bicycle model in the Cartesian coordinate frame, we describe the dynamics of a front-wheel driven and steered four-wheel vehicle with perfect rolling and no slipping \cite{chen2024aggfollower,han2023efficient}. The state vector is defined as $\mathbf{x} = (p_x, p_y, \theta, v, a_t, a_n, \phi, \kappa)^T$, where $\mathbf{p} = (p_x, p_y)^T$ represents the position at the centre of the rear wheels, $v$ is the longitudinal velocity w.r.t vehicle's body frame, $a_t$ and $a_n$ denote the longitudinal and lateral accelerations, $\phi$ is the steering angle of the front wheels, and $\kappa$ is the curvature. The complete trajectory representation $\sigma(t) : [0, T_s]$ is formulated as:
\begin{equation}
\sigma(t) = \sigma_i(t - \hat{T}_i), \forall i \in \{1, 2, ..., n\}, t \in [\hat{T}_i, \hat{T}_{i+1}),
\end{equation} where $T_s = \sum_{i=1}^n T_i$ is the duration of the entire trajectory, and $\hat{T}_i = \sum_{j=1}^{i-1} T_j$ is the timestamp of the starting point of the $i$-th segment, with $\hat{T}_1 = 0$. The comfort metric is defined as:
\begin{align}
C = \sum_{k=1}^3 \int_0^{T_k} \Big( & w_1 |a_t - a_t^*| + w_2 |a_n - a_n^*| \nonumber \\
& + w_3 |\dot{\phi} - \dot{\phi}^*| + w_4 |j_t - j_t^*| \nonumber \\
& + w_5 |j_n - j_n^*| + w_6 |\dot{\kappa} - \dot{\kappa}^*| \Big) dt,
\end{align} where $T_k \in \{1s, 2s, 3s\}$ represents the considered trajectory duration, $a_t^*$, $a_n^*$, $\dot{\phi}^*$, $j_t^*$, $j_n^*$, and $\dot{\kappa}^*$ are the corresponding values from the ground true trajectory, and $w_1, w_2, w_3, w_4, w_5, w_6$ are weighting factors for longitudinal acceleration, lateral acceleration, steering angle rate, longitudinal jerk, lateral jerk, and curvature rate, respectively. The longitudinal and lateral jerk, $j_t$ and $j_n$ are calculated as the time derivatives of $a_t$ and $a_n$, respectively.  


\section{Experiments}
\begin{figure*}[t]
  \centering
     \includegraphics[width=0.9\linewidth]{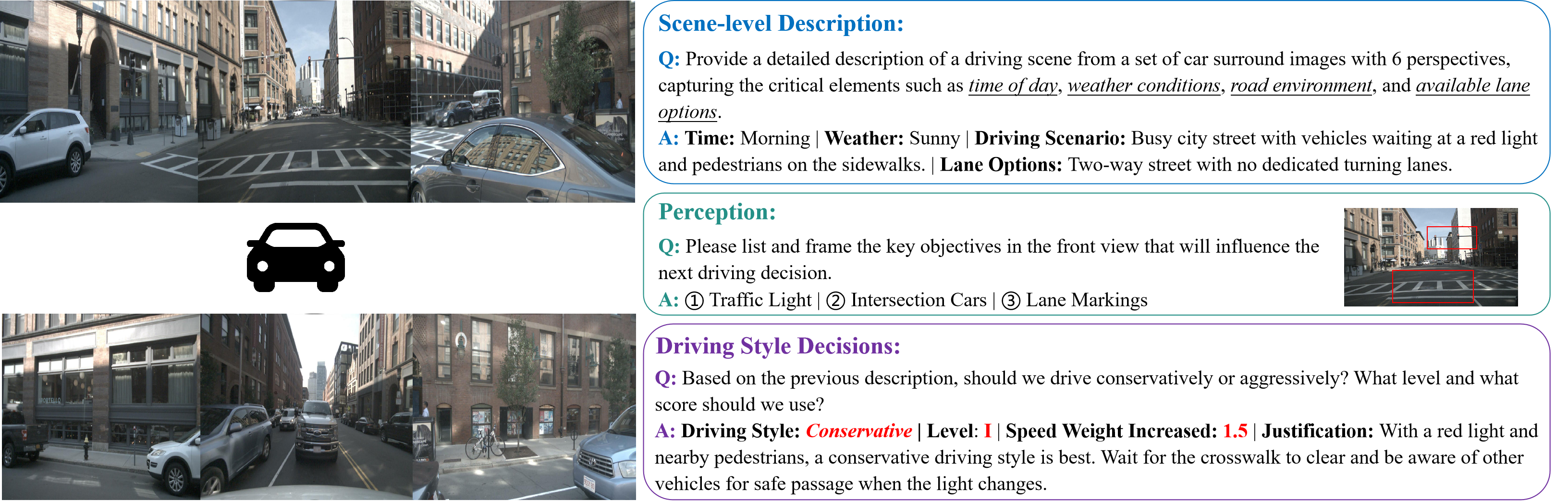}
   \caption{\small \textbf{Qualitative results of Llama 3.2V on nuScenes.} We show the questions (Q), context (C), and answers (A). Incorporating surround view images and textual data, the fine-tuning of driving styles via targeted weight modifications within the rule-based scorer. }
   \label{fig:driving_style_1}
\end{figure*}
\subsection{Experiment Setup}
\noindent\textbf{Datasets:} We evaluate ComDrive on two challenging datasets: nuScenes and real-world datasets. \textbf{nuScenes} \cite{nuscenes} comprises 1,000 driving scenes, each spanning 20 seconds. The \textbf{Real-World} dataset is our newly collected 500 hours of driving data, mainly collected in urban areas and highways, providing a variety of real-world driving scenarios.

\noindent\textbf{Metrics:} We employ a comprehensive evaluation encompassing performance, comfort, and efficiency metrics. Performance is assessed using L2 and collision metrics from SparseDrive \cite{sun2024sparsedrive}, while comfort is evaluated using our proposed metrics. Efficiency is measured by reporting FPS and GPU hours required for training. 

\noindent\textbf{Implementation Details:} ComDrive's training process involves multiple stages. We first train the sparse perception component following SparseDrive's approach \cite{sun2024sparsedrive}, resulting in ComDrive-S and ComDrive-B variants. The output then feeds into our diffusion-based motion planner for trajectory generation. We conduct end-to-end training of the entire ComDrive on 8 NVIDIA RTX 4090 GPUs, using AdamW optimizer \cite{loshchilov2017decoupled} with a weight decay of 0.01 and an initial learning rate of 5e-4.

\input{Tables/motion_planning_results}

\subsection{End-to-End Planning Results on the nuScenes}
\label{sec:nuScenes}
\begin{figure}[t]
  \centering
  \includegraphics[width=\linewidth]{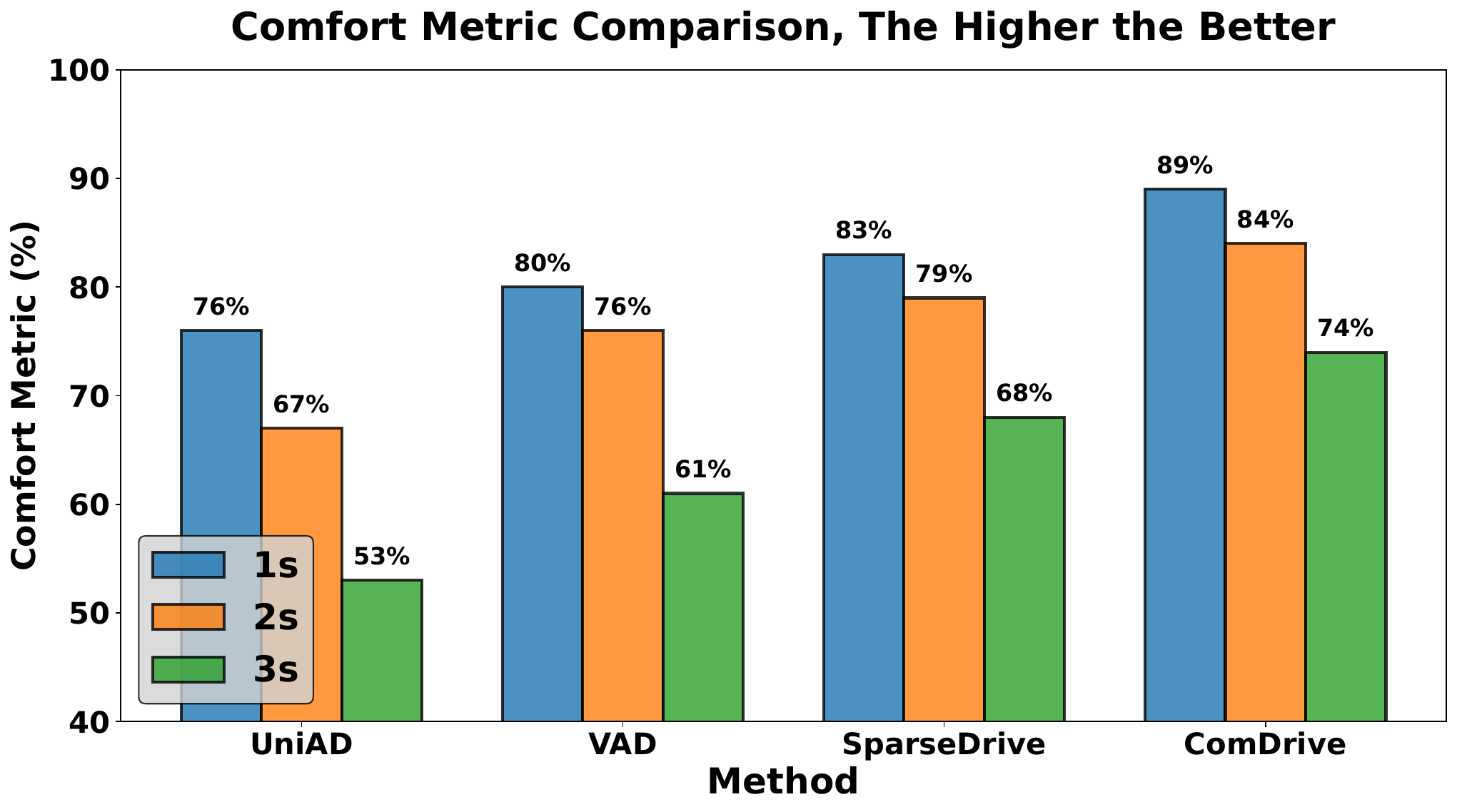}
  \caption{\small \textbf{Quantitative results of comfort metrics on the nuScenes dataset.}}
  \label{fig:nu-comfort}
\end{figure}

In Table \ref{tab:planning}, ComDrive outperforms previous Camera-based and LiDAR-based approaches in both performance and efficiency. ComDrive-S achieves a 17.8\% reduction in mean L2 error compared to UniAD while decreasing average collision rates by 68\%. This results from ComDrive's strong temporal consistency, as illustrated in Fig. \ref{fig:nuscene-1}. ComDrive-B, incorporating a stronger visual backbone \cite{sun2024sparsedrive} further reduces average L2 error and collision rates to 0.58 and 0.06, respectively. Notably, ComDrive-S operates at 16.1 FPS, 1.2x and 2.5x faster than SparseDrive and VAD, while achieving a 39.6\% improvement in 3s comfort level compared to UniAD (Fig. \ref{fig:nu-comfort}). The real-time DDIM generation during inference and the adaptive scorer's continuous selection of trajectories contribute significantly to the system's efficiency and comfort. Fig. \ref{fig:driving_style_1} demonstrates how Llama 3.2V multi-round dialogues enable efficient driving style adjustment. This capability allows the ego vehicle to adjust its driving approach based on environmental cues, such as distant traffic signals or road occupancy. The VLM's zero-shot reasoning enhances this utility. Upcoming ablation studies will further validate these findings. 

\begin{figure}[htp]
  \centering
     \includegraphics[width=\linewidth]{images/nu_results2.png}
   \caption{\small \textbf{Qualitative results on the nuScenes dataset.} Our ComDrive exhibits strong temporal consistency.}
   \label{fig:nuscene-1}
\end{figure}

\begin{figure}[htp]
  \centering
  \begin{subfigure}{0.5\linewidth}
    \centering
    \includegraphics[width=0.98\textwidth]{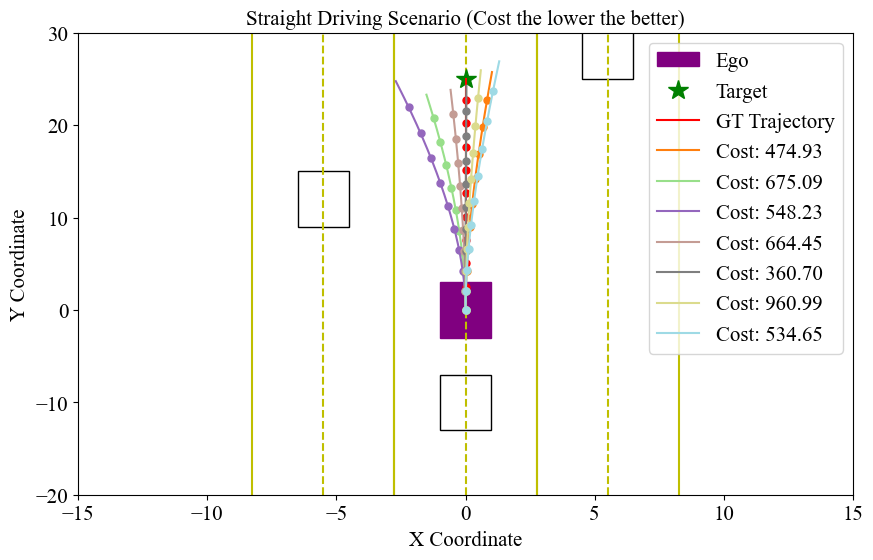}
    \caption{}
    \label{fig:real-1}
  \end{subfigure}%
  \hfill
  \begin{subfigure}{0.5\linewidth}
    \centering
    \includegraphics[width=0.95\textwidth]{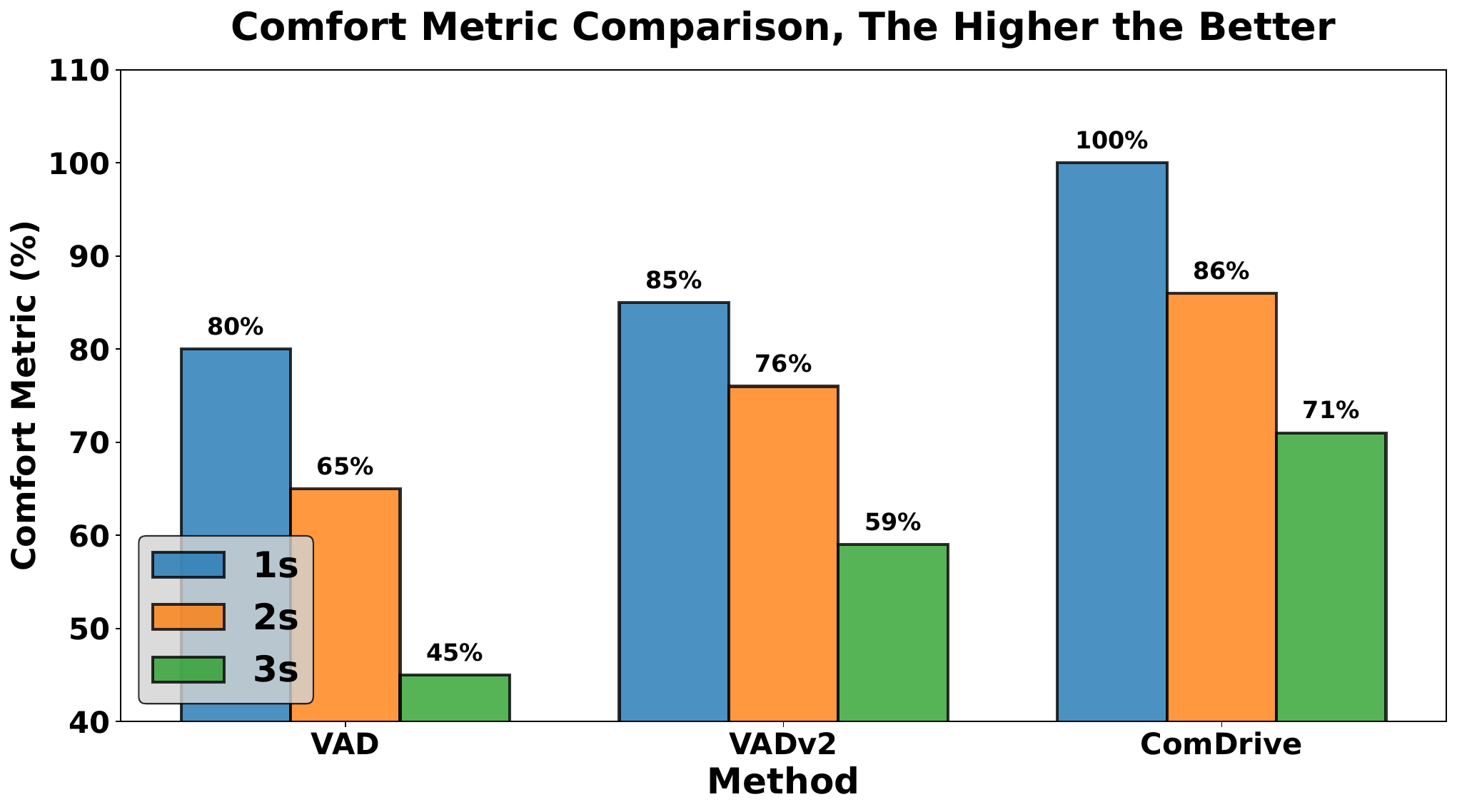}
    \caption{}
    \label{fig:comfort-index}
  \end{subfigure}%
  \hfill
 
  \caption{\small (a) showcase the trajectory generation and scoring process, with the optimal path, indicated by the grey trajectory in (a), being selected for vehicle control based on the lowest cost criterion. (b) shows the comparison results of ComDrive and two baselines in terms of the comfort metric in real-world data.}
  \label{fig:real_234}
\end{figure}

\subsection{Ablation Study on the nuScenes}

We conduct extensive experiments to study the effectiveness and necessity of each design choice proposed in our ComDrive. We use ComDrive-S as the default model for ablation.

\noindent\textbf{Trajectory Consistency:} Table \ref{tab:ablation_consistency} demonstrates the impact of various components on trajectory consistency. The full model achieves the lowest average L2 error and collision rate. Replacing DDIM with a Fixed Trajectory Set (FTS) significantly degrades performance, increasing L2 error by 53.3\%. This underscores DDIM's superiority in generating adaptive trajectories. Removing 3D representation, historical trajectory coordinates, or kinematic features (velocity, acceleration, yaw) all lead to performance drops, highlighting their importance in maintaining trajectory consistency.

\input{Tables/ablation_consistency}

\input{Tables/ablation_planning_mode}
\noindent\textbf{Vision-Language Model Comparison:} Table \ref{tab:ablation_combined} demonstrates that our rule-based approach achieves competitive L2 error (0.98) and collision rate (0.49\%), comparable to state-of-the-art models (e.g., GenAD). Integrating VLMs, especially Llama 3.2V, further improves these metrics while significantly enhancing comfort. Llama 3.2V achieves the best overall performance with an L2 error of 0.60, a collision rate of 0.07\%, and a comfort score of 74\%, representing improvements of 17.8\%, 66.7\%, and 19.3\% respectively compared to the rule-based approach. While different VLMs show similar safety performance, Llama 3.2V excels in comfort, highlighting that VLMs primarily fine-tune driving style for comfort, while our rule-based foundation maintains consistent safety standards.

\noindent\textbf{Anchor Points:} Our experiments show that 8 planning anchor points provide the optimal balance between accuracy and efficiency. This configuration achieves the lowest L2 error and collision rate while maintaining the highest comfort score (74\%). Increasing anchor points beyond 8 yields no further improvements, indicating that this number suffices for effective trajectory planning in most scenarios.

\noindent\textbf{DDIM Trajectory Averaging:} Table \ref{tab:ablation_combined} illustrates the crucial role of DDIM in our approach. Directly averaging the multi-modal trajectories generated by DDIM results in performance degradation across all metrics, due to the loss of essential multi-modal characteristics. In contrast, our refined DDIM with the VLM method significantly enhances performance, yielding a 7.7\% reduction in L2 error, a 22.2\% decrease in collision rate, and a notable 19.4\% improvement in comfort scores. These findings underscore the importance of preserving DDIM's multi-modal nature, enabling our model to generate diverse, context-appropriate trajectories that simultaneously enhance both safety metrics and passenger comfort.

\subsection{End-to-End Planning Results on the Real-World Dataset}
\label{sec:real-world}
The end-to-end planning results on the real-world dataset are shown in Fig. \ref{fig:real_234}. ComDrive generates consistent multimodal trajectories and selects the optimal path using our dual-stream scorer. Higher-cost trajectories (purple and green) deviate from the target or reduce comfort during turns, demonstrating our scorer's safety prioritization and interpretability. Fig. \ref{fig:real_234}b illustrates ComDrive's superior comfort metrics, with 1s trajectory segments reaching 100\% comfort, outperforming VAD by 20\%. The overall 3s trajectory comfort exceeds VADv2, highlighting our scorer's efficient lifelong evaluation capabilities. By adjusting the driving style through LIama 3.2V, the most comfortable straight trajectory is selected.

\section{Conclusion}

This paper presents ComDrive, a novel end-to-end autonomous driving system designed to address temporal consistency and passenger comfort challenges. Our approach integrates a sparse perception module for comprehensive 3D spatial representations, a diffusion-based motion planner for temporally consistent multi-modal trajectories, and a dual-stream adaptive trajectory scorer combining rule-based methods and VLMs to dynamically adjust driving styles. ComDrive enhances generalization and overall comfort. Experiments in both 2 datasets demonstrate its superior performance in generating temporally consistent and comfortable trajectories compared to state-of-the-art methods.

\bibliographystyle{IEEEtran}
\balance
\bibliography{root}

\end{document}

%% file: Tables/cost.tex
\begin{table}[!htb]
\centering
\small
\caption{Rule-based scorer initial weights (left) and driving style weights adjustment range (right)}
\setlength{\tabcolsep}{3pt}
\begin{tabular}{@{}c@{\hspace{4mm}}c@{}}
    \begin{tabular}{@{}lcr@{}}
    \toprule
    \textbf{Category} & \textbf{Cost} & \textbf{Weight} \\
    \midrule
    \multirow{4}{*}{Safety} 
        & \(w_{\text{coll}}\) & 5.0 \\
        & \(w_{\text{dev}}\) & 3.5 \\
        & \(w_{\text{dis}}\) & 1.5 \\
        & \(w_{\text{speed}}\) & 2.5 \\
    \midrule
    \multirow{3}{*}{Comfort} 
        & \(w_{\text{lat}}\) & 1.5 \\
        & \(w_{\text{lon}}\) & 4.5 \\
        & \(w_{\text{cent}}\) & 3.0 \\
    \bottomrule
    \end{tabular}
    &
    \begin{tabular}{@{}lcc@{}}
    \toprule
    \textbf{Style} & \textbf{Level} & \textbf{Range} \\
    \midrule
    \multirow{4}{*}{Aggressive} 
        & I   & 1.5 - 3.0 \\
        & II  & 1.0 - 1.4 \\
        & III & 0.1 - 0.9 \\
        & & \\
    \midrule
    \multirow{3}{*}{Conservative} 
        & I   & 1.5 - 3.0 \\
        & II  & 1.0 - 1.4 \\
        & III & 0.1 - 0.9 \\
    \bottomrule
    \end{tabular}
\end{tabular}

\label{tab:combined_weights}
\end{table}

%% file: Tables/motion_planning_results.tex
\definecolor{Gray}{gray}{0.85} 
\begin{table}[t]
\centering
\caption{Planning results on the nuScenes validation dataset.  $\dagger$: Reproduced with official checkpoint. }
\vspace{5pt}
\footnotesize
\setlength{\tabcolsep}{0.8mm}
\resizebox{\columnwidth}{!}{
\begin{tabular}{l|c|c|cccc|cccc|c}
\toprule
\multirow{2}{*}{Method} & \multirow{2}{*}{Input} & \multirow{2}{*}{Reference} & \multicolumn{4}{c|}{L2($m$)$ \downarrow$} & \multicolumn{4}{c|}{Collision Rate(\%)$ \downarrow$} & \multirow{2}{*}{FPS $\uparrow$} \\
& & & 1$s$ & 2$s$ & 3$s$ & \cellcolor{Gray}Avg. & 1$s$ & 2$s$ & 3$s$ & \cellcolor{Gray}Avg.\\
\midrule
IL~\cite{IL} & LiDAR & ICML 2006 & 0.44 & 1.15 & 2.47 & \cellcolor{Gray}1.35 & 0.08 & 0.27 & 1.95 & \cellcolor{Gray}0.77 & \cellcolor{Gray}-\\
FF~\cite{ff} & LiDAR & CVPR 2021 & 0.55 & 1.20 & 2.54 & \cellcolor{Gray}1.43 & 0.06 & 0.17 & 1.07 & \cellcolor{Gray}0.43 & \cellcolor{Gray}-\\
EO~\cite{eo} & LiDAR & ECCV 2022 & 0.67 & 1.36 & 2.78 & \cellcolor{Gray}1.60 & 0.04 & 0.09 & 0.88 & \cellcolor{Gray}0.33 & \cellcolor{Gray}-\\
\midrule
ST-P3~\cite{stp3} & Camera & ECCV 2022 & 1.33 & 2.11 & 2.90 & \cellcolor{Gray}2.11 & 0.23 & 0.62 & 1.27 & \cellcolor{Gray}0.71 & \cellcolor{Gray}1.6\\
OccNet \cite{tong2023scene} & Camera & ICCV 2023  & 1.29 & 2.13 & 2.99 & \cellcolor{Gray}2.14 & 0.21 & 0.59 & 1.37  & \cellcolor{Gray}0.72 & \cellcolor{Gray}2.6\\ 
UniAD$^\dagger$~\cite{uniad} & Camera & CVPR 2023 & 0.45 & 0.70 & 1.04 & \cellcolor{Gray}0.73 & 0.62 & 0.58 & 0.63  &\cellcolor{Gray} 0.61 & \cellcolor{Gray}1.8\\
VAD$^\dagger$~\cite{vad} & Camera & ICCV 2023 & 0.41 & 0.70 & 1.05 & \cellcolor{Gray}0.72 & 0.03 & 0.19 & 0.43  & \cellcolor{Gray}0.21 & \cellcolor{Gray}4.5\\ 
SparseDrive \cite{sun2024sparsedrive} & Camera & arXiv 2024 & \textbf{0.29} & \textbf{0.58} & 0.96 & \cellcolor{Gray}0.61 & \textbf{0.01} & \textbf{0.05} & 0.18 & \cellcolor{Gray}0.08 & \cellcolor{Gray}9.0\\
OccWorld-T \cite{zheng2023occworld} & Camera & ECCV 2024 & 0.54 & 1.36 & 2.66 & \cellcolor{Gray}1.52 & 0.12 & 0.40 & 1.59  & \cellcolor{Gray}0.70 & \cellcolor{Gray}2.8\\
OccWorld-S \cite{zheng2023occworld} & Camera & ECCV 2024 & 0.67 & 1.69 & 3.13 & \cellcolor{Gray}1.83 & 0.19 & 1.28 &  4.59 & \cellcolor{Gray}2.02 & \cellcolor{Gray}2.8\\
GenAD \cite{zheng2024genad} & Camera & ECCV 2024 & 0.36 & 0.83 & 1.55 & \cellcolor{Gray}0.91 & 0.06 & 0.23 & 1.00 & \cellcolor{Gray}0.43 & \cellcolor{Gray}6.7\\

\midrule
ComDrive-S (Ours) & Camera & - & 0.31 & \textbf{0.58} & \textbf{0.93} & \cellcolor{Gray} \textbf{0.60} & \textbf{0.01} & \textbf{0.05} & \textbf{0.16} & \cellcolor{Gray}\textbf{0.07} & \cellcolor{Gray}\textbf{16.1}\\
ComDrive-B (Ours) & Camera & - & 0.30 & \textbf{0.56} & \textbf{0.89} & \cellcolor{Gray} \textbf{0.58} & \textbf{0.00} & \textbf{0.03} & \textbf{0.14} & \cellcolor{Gray}\textbf{0.06} & \cellcolor{Gray}\textbf{10.0}\\
\bottomrule
\end{tabular}
}
\label{tab:planning}
\end{table}

%% file: Tables/ablation_consistency.tex
\begin{table}[t]
\centering
\caption{Ablation study on trajectory consistency. "DDIM" denotes the use of DDIM Model; "3D" represents the incorporation of 3D representation; "HTC" refers to the inclusion of Historical Trajectory Coordinates; "KF" signifies the use of Kinematic Features (velocity, acceleration, yaw); "FTS" indicates the use of a Fixed Trajectory Set.}
\label{tab:ablation_consistency}
\vspace{5pt}
\scriptsize
\setlength{\tabcolsep}{1.0mm}
\begin{tabular}{ccccc|cc}
\toprule
\multirow{2}{*}{DDIM} &
\multirow{2}{*}{3D} &
\multirow{2}{*}{HTC} &
\multirow{2}{*}{KF} &
\multirow{2}{*}{FTS} &
\multicolumn{1}{c}{L2(m)} &
\multicolumn{1}{c}{Coll.(\%)} \\
& & & & & Avg. & Avg.\\
\midrule
\checkmark & \checkmark & \checkmark & \checkmark & & \cellcolor{gray!20}\textbf{0.60} & \cellcolor{gray!20}\textbf{0.07} \\
& & & & \checkmark & \cellcolor{gray!20}0.92 & \cellcolor{gray!20}0.30 \\
\checkmark & & \checkmark & \checkmark & & \cellcolor{gray!20}0.68 & \cellcolor{gray!20}0.11 \\
\checkmark & \checkmark & & \checkmark & & \cellcolor{gray!20}0.76 & \cellcolor{gray!20}0.25 \\
\checkmark & \checkmark & \checkmark & & & \cellcolor{gray!20}0.71 & \cellcolor{gray!20}0.11 \\
\bottomrule
\end{tabular}
\end{table}

%% file: Tables/ablation_planning_mode.tex
\begin{table}[t]
\centering
\caption{Ablation studies on VLM comparison, anchor points, and DDIM trajectory averaging.}
\label{tab:ablation_combined}
\vspace{5pt}
 \scriptsize
\setlength{\tabcolsep}{1.2mm}
\begin{tabular}{l|c|c|c}
\toprule
Model/\#Anchors/Method & L2(m) & Coll.(\%) & Comfort \\
 & Avg. & Avg. & (\%)  \\
\midrule
\multicolumn{4}{c}{VLM Comparison} \\
\midrule
Rule-based (ours) & 0.98 & 0.49 & 62 \\
GPT-4o \cite{achiam2023gpt} & 0.63 & 0.10 & 71 \\
Qwen2-VL \cite{Qwen2VL} & 0.63 & 0.09 & 72 \\
\cellcolor{gray!20}Llama 3.2V & \cellcolor{gray!20}\textbf{0.60} & \cellcolor{gray!20}\textbf{0.07} & \cellcolor{gray!20}\textbf{74} \\
\midrule
\multicolumn{4}{c}{Anchor Points} \\
\midrule
4 & 0.70 & 0.18 & 73 \\
6 & 0.66 & 0.14 & 73 \\
\cellcolor{gray!20}8 & \cellcolor{gray!20}\textbf{0.60} & \cellcolor{gray!20}\textbf{0.06} & \cellcolor{gray!20}\textbf{74} \\
10 & 0.68 & 0.16 & \cellcolor{gray!20}\textbf{74} \\
\midrule
\multicolumn{4}{c}{DDIM Trajectory Averaging} \\
\midrule
DDIM Avg. w/ VLM & 0.65 & 0.09 & 62 \\
\cellcolor{gray!20}DDIM w/ VLM & \cellcolor{gray!20}\textbf{0.60} & \cellcolor{gray!20}\textbf{0.07} & \cellcolor{gray!20}\textbf{74} \\
\bottomrule
\end{tabular}
\end{table}